\documentclass[runningheads]{llncs}

\usepackage{iciapabbrv}

\usepackage{graphicx}
\usepackage{booktabs}

\usepackage[accsupp]{axessibility}  

\usepackage{hyperref}

\usepackage{orcidlink}

\begin{document}

\title{Predicting Soccer Penalty Kick Direction Using Human Action Recognition} 

\titlerunning{Predicting Soccer Penalty Kick Direction Using HAR and NNs}

\author{David Freire-Obreg\'on\inst{1}\orcidlink{0000-0003-2378-4277} \and
Oliverio J. Santana\orcidlink{0000-0001-7511-5783} \and
Javier Lorenzo-Navarro\inst{1}\orcidlink{0000-0002-2834-2067} \and
Daniel Hernández-Sosa\orcidlink{0000-0003-3022-7698} \and
Modesto Castrill\'on-Santana\inst{1}\orcidlink{0000-0002-8673-2725}}

\authorrunning{D. Freire-Obregón et al.}

\institute{SIANI, Universidad de Las Palmas de Gran Canaria, Spain\\
\email{david.freire@ulpgc.es} 
}
\maketitle

\begin{abstract}
Action anticipation has become a prominent topic in Human Action Recognition (HAR). However, its application to real-world sports scenarios remains limited by the availability of suitable annotated datasets. This work presents a novel dataset of manually annotated soccer penalty kicks to predict shot direction based on pre-kick player movements. We propose a deep learning classifier to benchmark this dataset that integrates HAR-based feature embeddings with contextual metadata. We evaluate twenty-two backbone models across seven architecture families (MViTv2, MViTv1, SlowFast, Slow, X3D, I3D, C2D), achieving up to 63.9\% accuracy in predicting shot direction (left or right)—outperforming the real goalkeepers' decisions. These results demonstrate the dataset's value for anticipatory action recognition and validate our model's potential as a generalizable approach for sports-based predictive tasks.

\keywords{Human Action Recognition\and Action Prediction \and Soccer \and Vision Transformers \and Penalty Kick.}
\end{abstract}

\section{Introduction}

Human action recognition (HAR) in video has gained significant attention in recent years \cite{NEW1}, with action anticipation—predicting future actions from past observations—being one of its most challenging tasks \cite{NEW3}. Deep learning approaches are increasingly used for this purpose \cite{NEW4}, but they require specialized datasets. In FIFA World Cups, shootouts decided three titles (1982, 1994, 2006), while UEFA Champions League finals saw penalties settle 30\% of matches over 23 years. These moments highlight their significance in club and international soccer. This paper presents a dataset of 1,010 penalty kick clips from official soccer matches. It aims to train neural networks with HAR embeddings to predict shot direction. Model development is based on a representative subset of the data. Penalty kicks are pivotal in soccer, shaping match outcomes and tournament histories.

\begin{figure}[t]  
    \centering
    \includegraphics[scale=0.25]{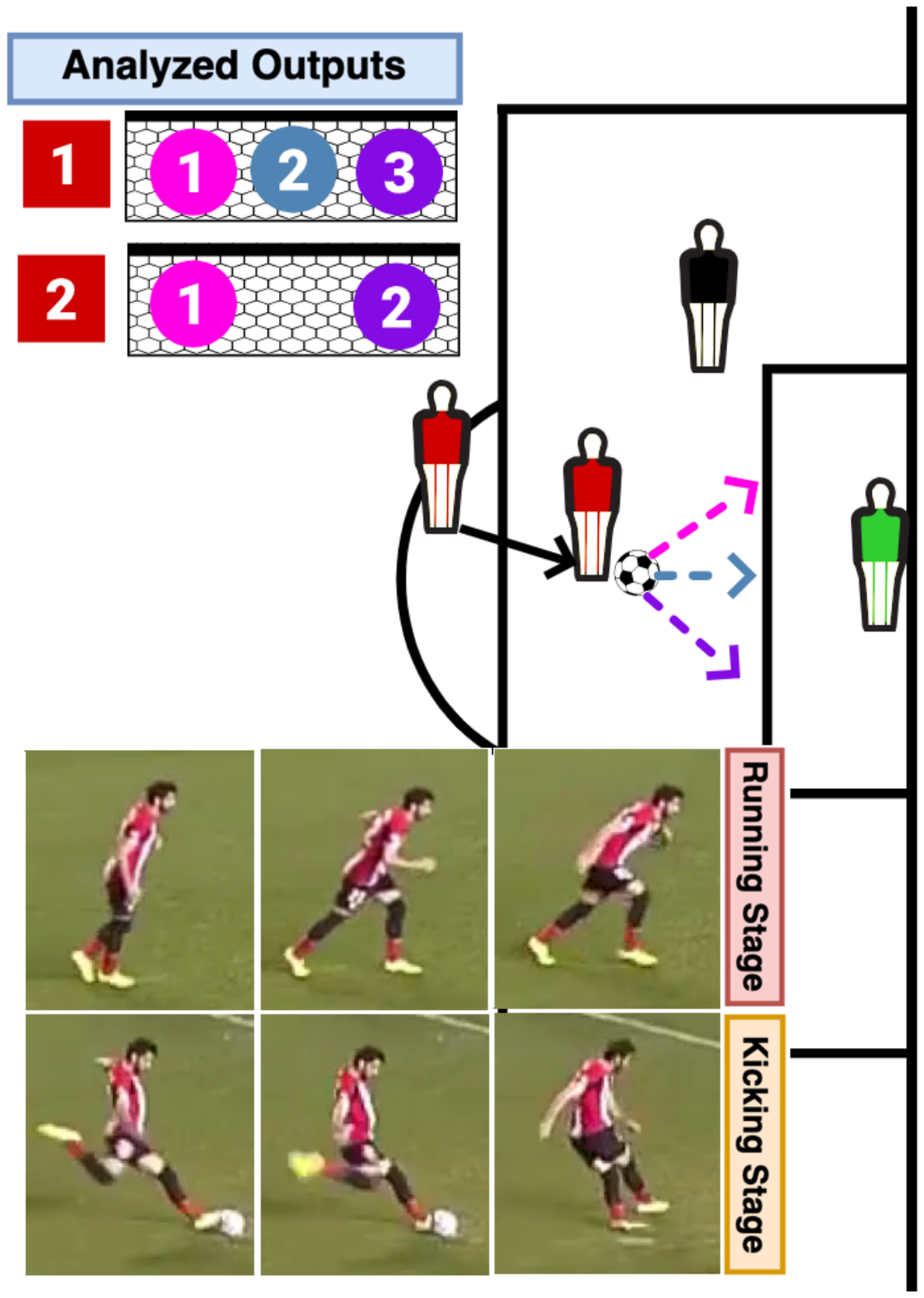}
    \caption{\textbf{Penalty-kick direction estimation.} Our method predicts ball direction by combining metadata and HAR embeddings, using only kicker actions without ball trajectory data.}
    \label{fig:intro_img}
\end{figure}

Integrating technology into soccer has significantly improved the analysis of game dynamics. The combination of video processing, data visualization, and automation enables more profound insights into player behavior \cite{He22}, tactical formations \cite{Li23}, and key moment detection \cite{Deliege20}. These advances support applications such as strategy analysis and predictive modeling. In this context, penalty kick outcome prediction can benefit from incorporating factors like tempo, technique, and individual skills.

The motivation behind this work is to support goalkeeper decision-making through data-driven video analysis, both for offline tactical preparation and potentially for online support during matches, assuming future integration with a real-time tracking system. While personalized models could be valuable, our approach focuses on general motion cues that may reveal intent in real-time. The proposed method processes video clips using HAR encoders split into running and kicking phases (see Figure \ref{fig:intro_img}), combined with contextual metadata in a two-stream classifier. We evaluate three-class (left, center, right) and two-class (left, right) formulations. The binary setup helps mitigate ambiguity in the "center" class, which can be visually and semantically unclear due to inconsistencies in human annotation and variations in camera angles. Results show that shot direction can be predicted with over 50\% accuracy—exceeding 60\% in the binary setting—outperforming the behaviors observed in real goalkeepers.

\section{Related work}
\label{sec:relatedwork}

Sports analysis has gained significant research interest, with technology transforming our understanding of sports and athlete performance. This research focuses on sports datasets and associated challenges, specifically within the context of ball sports, where the interaction between players and the ball is crucial in determining outcomes.

Sports datasets are generally categorized into still-image and video-sequence types. Still-image datasets, like the Leeds Sports Pose Dataset, focus on image classification and athlete poses. Video-sequence datasets, like the MTL-AQA diving dataset \cite{Parmar19} and Fis-V skating dataset \cite{Xu20}, provide dynamic scene insights for action recognition and player tracking, even under challenging conditions. Usually, the semantic structure of sports video content is organized into four layers: raw video, object, event, and semantic layers \cite{Shih18}. Raw video is processed to identify objects like players \cite{Tianxiao20} and the ball \cite{Shaobo19}. The event layer captures actions and interactions, with research areas including action recognition \cite{freire22icpr}, re-identification \cite{Akan23}, facial expression recognition \cite{ojsantana22mtool}, and trajectory prediction \cite{Teranishi20}. The semantic layer summarizes the footage's content \cite{Cioppa18}. For this study, the semantic layer categorizes penalty kick outcomes based on the kicker's actions \cite{Artilesicpram24}, excluding team dynamics to focus on individual actions and minimizing noise from other individuals present during the shot.

The goalkeeper's behavior during penalty kicks has been widely studied due to its key role in high-pressure, structured situations. Hunter et al. \cite{Hunter2022} proposed a model to improve goal likelihood by analyzing kicker and goalkeeper strategies. Earlier work \cite{Hunter2018} showed that goalkeepers anticipate shot direction using the kicker's cues, with kick speed being critical. Noël et al. \cite{Noel2015} examined two strategies: keeper-independent (pre-decided shot direction) and keeper-dependent (reactive to the goalkeeper), finding the former more common (78–86\%) with similar success rates. Recent work highlights goalkeeper deception, showing that subtle misleading actions can reduce goal conversion, especially when kickers attend to the goalkeeper \cite{Zheng02072024}. Our work proposes an automated approach, designing a deep learning model to predict the shot direction from kicker movements.

\section{Method}
\label{sec:formulation}

\subsection{Problem Formulation}
Given a set of $m$ of penalty kick samples, an observation $o^{(i)}=(F^{(i)},y^{(i)},\Gamma^{(i)})$ for $i=1 \dots m$, is composed of annotated footage $F^{(i)}$, a label $y^{(i)}$ that corresponds to the direction of the shot, and contextual information $\Gamma^{(i)}$. The latter includes two distinct input variables known before the kicking action: the penalty-kick field side and the kicking foot. Both variables serve as binary indicators, distinguishing between left and right. Regarding the labels, two different scenarios are considered in this work: $y^{(i)} \in [left, center, right]$ and $y^{(i)} \in [left, right]$. Thus, the problem can be stated as finding the model parameter set $\theta$ of an end-to-end classifier that minimizes the log loss function:
\begin{equation}
J(\theta) = -\frac{1}{m} \sum_{i=1}^{m} \sum_{k=1}^{n} p(y^{(i)}=k) \log(p(\hat{y}{(i)}=k))
\label{eq:minim}
\end{equation}
where $m$ is the number of penalty kick samples, $n$ is the number of classes, $p(y^{(i)}=k)$ is the probability that the observation $o^{(i)}$ belongs to class $k$, and $p(\hat{y}^{(i)}=k)$ is the probability that the predicted class for the observation $o^{(i)}$ belongs to class $k$.

\subsection{Multistage and Classifier Formulation}

First, the original footage $F^{(i)}$ undergoes a pre-processing step to constrain the context, resulting in $F'^{(i)}$. Then, each footage $F'^{(i)}$ is manually divided into two sets of frames: 32 for the running stage and 16 for the kicking stage, values empirically chosen to cover typical pre-kick motions and maintain temporal balance across clips. Consequently, $F'^{(i)} = F'^{(i)}_{run} \cup F'^{(i)}_{kick}$.

Secondly, the feature extraction process involves passing the frames from the running and kicking stages through HAR backbones, grouped into chunks of frames, denoted as $\Omega(F'^{(i)}_{run})=\{f^{i,1}_{run}, \dots, f^{i,n^i_f}_{run}\}$ and $\Omega(F'^{(i)}_{kick})=\{f^{i,1}_{kick}, \dots, f^{i,n^i_f}_{kick}\}$ respectively, where $f^{i,j}_{run}$ and $f^{i,j}_{kick}$ represent the embedding of the $j$-th chunk of the footage $F'^{(i)}$, being $n^i_f$ the number of chunks into which the footage was split. In order to obtain a single embedding per footage, a pooling operation is introduced as follows:
\begin{equation}
\begin{aligned}
T^{(i)}_{run} &= pool(\Omega(F_{run}^{'(i)})) \\
T^{(i)}_{kick} &= pool(\Omega(F_{kick}^{'(i)}))
\label{eq:pooling}
\end{aligned}
\end{equation}
where $pool \in [Average, Max]$ denotes the pooling operation. The footage embeddings, denoted as $T^{(i)}_{run}$ and $T^{(i)}_{kick}$, are then utilized as input for a subsequent classifier.

Consequently, if $m$ represents the total number of penalty-kick instances in the dataset, the objective is to learn a mapping function $C$ that predicts the kick direction based on the features extracted from the running and kicking stages, and contextual information:
\begin{equation}
C: (T_{\text{run}}^{(i)}, T_{\text{kick}}^{(i)}, \Gamma^{(i)}) \rightarrow y^{(i)}
\end{equation}
where $T_{\text{run}}^{(i)}$ and $T_{\text{kick}}^{(i)}$ are the embeddings corresponding to the $i$-th penalty-kick instance, and $y^{(i)}$ is the direction of the shot, as previously defined. The classifier, $C$, integrates information from both the running and kicking stages along with contextual information ($\Gamma^{(i)}$).

\begin{figure}[t]  
    \centering
    \includegraphics[scale=0.53]{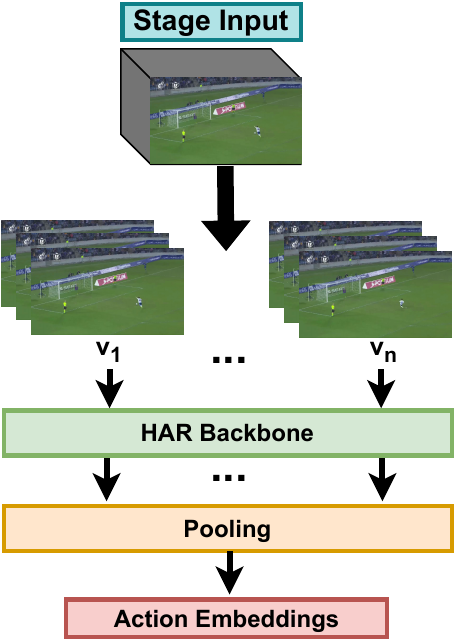}
    \caption{\textbf{Embeddings extraction module.} The video is downsampled into $n$ clips. A pre-trained HAR model extracts features, combined via pooling (average or max) to produce a final tensor for the classifier.}
    \label{fig:embeddings}
\end{figure}

\section{Pipeline description}
\label{sec:pipeline}

\begin{figure*}[t]  
    \centering
    \includegraphics[scale=0.55]{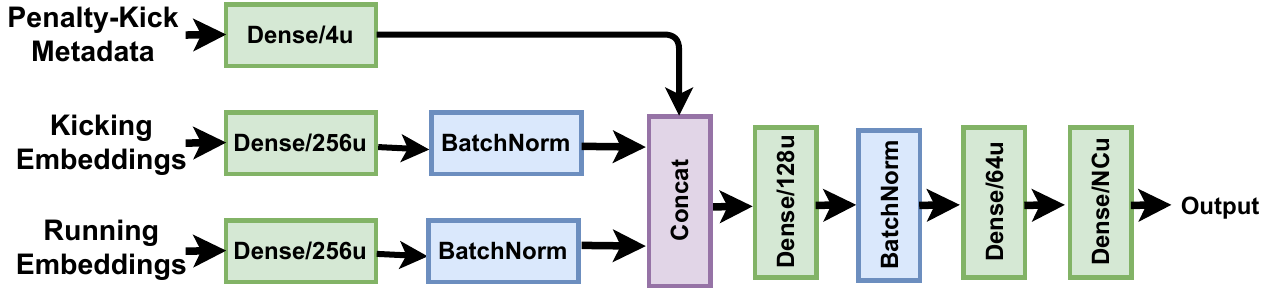}
    \caption{\textbf{The proposed classifier.} It combines HAR backbone features and penalty-kick metadata for enhanced decision-making during running and kicking stages.}
    \label{fig:classifier}
\end{figure*}

This paper introduces a sequential pipeline with two core modules: embedding extraction and a classifier. Video pre-processing is manually executed before the pipeline. Visual representations are shown in Figures \ref{fig:embeddings} and \ref{fig:classifier}.

\textbf{Step 1: Video Pre-processing: Context Constraint}. To optimize embedding quality, extraneous elements (e.g., unrelated players, staff, supporters, referees) are removed from the input footage~\cite{freire22icpr}. The primary subject, the kicker, is isolated using ByteTrack~\cite{zhang2021bytetrack}, a precise multi-object tracking network. Context-constrained pre-processing creates an ideal experimental setting.

For a kicker ($i$) at time ($t$) within interval ($[0, T]$), the bounding box ($BB^{(i)}(t)$) outlines the kicker's area in the frame. The pre-processed frame ($F'^{(i)}(t)$) is generated by superimposing $BB^{(i)}(t)$ onto the average frame ($\overline{f}$), computed from $\tau$ frames, where $\tau$ corresponds to the total number of frames in the considered sequence (e.g., $\tau = 48$ for a sequence composed of 32 running and 16 kicking frames):

\begin{equation}
F'^{(i)} = BB^{(i)}(t) \cup \overline{f}
\end{equation}

The $\cup$ operation aligns the bounding box with the static background, producing footage with the kicker as the sole moving element. 
We first compute an average frame over the entire input sequence to represent the static background to achieve this. 
Then, for each frame in the sequence, the kicker's bounding box is extracted and superimposed onto this average frame. 
This results in a new sequence where the background remains fixed and only the kicker's motion is preserved, allowing the model to focus on the relevant dynamics. Since there is no camera movement during the analyzed stage of the penalty kick, this alignment is direct and does not require additional spatial transformations. Moreover, the footage was temporally cropped to begin at the start of the penalty sequence. In cases where the running phase is shorter than expected (i.e., 32 frames), temporal padding is applied to maintain a consistent input length.

\textbf{Step 2: HAR Embeddings Extraction}. Input footage $F'^{(i)}$ is downsampled into $n^{i}_f$ overlapping clips, spaced one frame apart. A pre-trained HAR model extracts features consolidated via pooling (equation \ref{eq:pooling}) into a single embedding. Two pooling methods (average and max) are evaluated. Twenty-two HAR backbones, ranging from MViT and Slowfast to C2D and X3D, are tested for kick direction challenges. The C2D model uses 2D CNNs for spatial feature extraction \cite{C2D14}, while SlowFast employs a dual-pathway approach for temporal resolution \cite{SlowFast19}. Enhanced versions, SlowFastNLN and SlowNLN, integrate non-local operations for spatiotemporal dependencies \cite{NonlocalNN17} (NLN). Contrary, the I3D model processes video clips as 3D volumes \cite{Carreira17}, and its variant I3D\_NLN includes non-local operations. The X3D model family (X3D-XS to X3D-L) adapts spatial and temporal dimensions \cite{Feichtenhofer20}. Vision Transformers like MViTv1 \cite{Fan21} and MViTv2 \cite{Li21} use multiscale feature hierarchies for improved recognition.

\textbf{Step 3: Classification Module}. The classifier integrates HAR backbone features and metadata, including the pitch side and the kicker's foot (Figure \ref{fig:classifier}). Inputs are processed via fully connected layers, concatenated, and passed through additional layers. Outputs classify ball position as \textit{left}, \textit{right}, or \textit{center}, or as a binary \textit{left/right} decision.

\begin{figure*}[t]  
    \centering
    \includegraphics[scale=0.6]{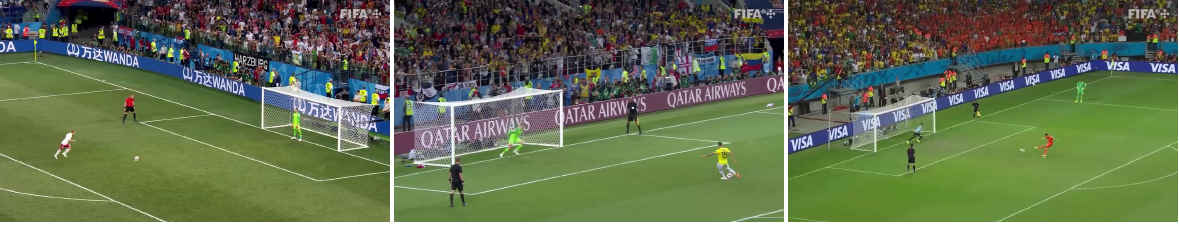}
    \caption{\textbf{Penalty-kick dataset samples.} The dataset, sourced from the Internet, exhibits pose, scale, and lighting variability. Clips are edited to focus on the running and kicking stages.}
    \label{fig:dataset}
\end{figure*}

\section{Experiments and Results} 
\label{sec:experiments}

This section covers dataset acquisition, experimental setup, and results. It begins with dataset details, including acquisition and cleaning, followed by methodology insights and data partitioning. Finally, it summarizes the experimental outcomes.

\textbf{Dataset}. No publicly available dataset for soccer penalty kicks exists. Our dataset, collected from online sources, exhibits variations in pose, scale, and lighting (see Figure~\ref{fig:dataset}). Web scraping with keywords like ``penalty-kick shootout'' yielded 1010 clips, trimmed to include the run-up and kick result. The original videos are in $1280\times720$ resolution and last between 3 and 6 seconds. They capture the player from just before starting the run-up until the ball reaches the net, the goalkeeper, the post, or goes out of play. For our study, each clip was trimmed to 48 frames: 32 frames capturing the player's run-up, followed by 8 frames before and 8 frames after the kicking moment. This interval excludes visual information about the ball's trajectory or outcome, focusing solely on kick motion cues.

Camera viewpoint and video duration were critical. Clips with unsuitable perspectives (e.g., behind the goalkeeper) or fewer than 64 frames were excluded, reducing the dataset to 640 clips. ByteTrack ensured acceptable detection quality. Metadata, such as the pitch side and the kicking foot, was incorporated to enrich the dataset. While contextual features such as the kicking foot are manually annotated in this study, they could be inferred automatically in real-world settings using pose estimation or approach-angle analysis. Of the 640 clips, 395 penalties were on the right side of the pitch, 245 on the left, with 498 taken by right-footed players and 142 by left-footed players.

\begin{table*}[!htbp]
 \centering
 \begin{tabular}{|l|c|c|c|c|c|c|c|}
 \hline
  \small Architecture & \small Best Model & \small \#Frames &\small Pooling &\small Accuracy & \small Precision & \small Recall & \small F1-Score\\
 \hline
   \hline 
 \footnotesize GK Baseline &\footnotesize - & \footnotesize -  & \footnotesize - &\footnotesize 46.0\% &\footnotesize 35.3\% &\footnotesize 44.2\% &\footnotesize 38.4\% \\
 \hline 
 \footnotesize C2D~\cite{C2D14} &\footnotesize C2D\_R50 & \footnotesize 8  & \footnotesize Max &\footnotesize 47.1\% &\footnotesize 33.1\% &\footnotesize 45.6\% &\footnotesize 38.4\% \\ 
 \hline
 
 \footnotesize I3D~\cite{Carreira17} &\footnotesize I3D\_R50 & \footnotesize 8  &\footnotesize Average & \footnotesize 45.0\% & \footnotesize 32.7\% & \footnotesize 46.3\% & \footnotesize 38.3\% \\
 \hline

\footnotesize Slow~\cite{Slow21} & \footnotesize Slow8x8 &\footnotesize  8  &\footnotesize Average & \footnotesize  46.7\% & \footnotesize  28.9\% & \footnotesize  45.9\% & \footnotesize  35.5\% \\
 \hline
 
 \footnotesize SlowFast~\cite{SlowFast19} & \footnotesize SlowFast4x16 & \footnotesize 32  &\footnotesize Average & \footnotesize 46.2\% & \footnotesize 33.6\% & \footnotesize 46.4\% & \footnotesize 39.1\% \\
 \hline
 
 \footnotesize NLN~\cite{NonlocalNN17}  &\footnotesize Slow\_NLN\_4x16 &\footnotesize 32  &\footnotesize Max & \footnotesize  45.1\% & \footnotesize  32.8\% & \footnotesize  47.8\% & \footnotesize  38.9\% \\
 \hline
  \footnotesize X3D~\cite{Feichtenhofer20} &\footnotesize X3D\_M &\footnotesize 13 &\footnotesize Max & \footnotesize  45.9\% & \footnotesize  33.7\% & \footnotesize 45.3\% & \footnotesize  38.6\% \\
  \hline
  
  \footnotesize MViTv1~\cite{Fan21} &\footnotesize MViT\_CONV &\footnotesize 32 &\footnotesize Average & \footnotesize  \textbf{51.9\%} & \footnotesize  34.4\% & \footnotesize \textbf{47.1\%} & \footnotesize  39.7\% \\
  \hline

  \footnotesize MViTv2~\cite{Li21} &\footnotesize MViTv2\_S &\footnotesize 16 &\footnotesize Average & \footnotesize  51.6\% & \footnotesize  \textbf{35.5\%} & \footnotesize 45.8\% & \footnotesize  \textbf{40.1\%} \\
 \hline
 \end{tabular}
\caption{\textbf{Performance Comparison of HAR Architectures for Kick Estimation with Three Classes}. This table compares the top backbones used in penalty-kick direction estimation across different architectures, along with the goalkeeper baseline (GK Baseline), corresponding to the actual direction chosen by the goalkeeper during each penalty kick. The table lists the architectures, optimal model variants, number of frames used for HAR embedding extraction, and additional performance metrics.}
  \label{tab:3cls_shoot_zone}
 \end{table*}


\textbf{Experimental Setup}. We employed a 10-fold cross-validation strategy on our 640-sample dataset, with eight folds used for training, one for validation, and one for testing. Each fold contains 64 samples, roughly preserving the class distribution: $\sim$30 right (47.3\%), $\sim$23 left (35.8\%), and $\sim$11 center (16.9\%). Although class imbalance is present, we compensated for it by applying class weights during training. Underrepresented classes, such as \textit{center}, were given higher importance in the loss function, helping the model learn from all classes effectively. While each fold includes 64 samples for testing and validation, cross-validation over 10 folds provides comprehensive coverage across the whole dataset, thereby mitigating the risks of bias and enhancing robustness. The dataset consists of 229 left, 303 right, and 108 center-directed penalty kicks, showing class imbalance.

Each architecture in Tables \ref{tab:3cls_shoot_zone} and \ref{tab:2cls_shoot_zone} includes the best-performing instance among several evaluated variants. The column \#Frames indicates the number of frames processed per HAR input clip; shorter clips were temporally padded at the start. We evaluated average and max pooling strategies for embedding aggregation and selected the method that achieved the highest average validation performance during cross-validation.

\begin{table*}[!htbp]
 \centering
 \begin{tabular}{|l|c|c|c|c|c|c|c|}
 \hline
  \small Architecture & \small Best Model & \small \#Frames &\small Pooling &\small Accuracy & \small Precision & \small Recall & \small F1-Score\\
 \hline
  \hline 
 \footnotesize GK Baseline &\footnotesize - & \footnotesize -  & \footnotesize - &\footnotesize 54.2\% &\footnotesize 53.3\% &\footnotesize 53.2\% &\footnotesize 53.3\% \\
 \hline 
 \footnotesize C2D~\cite{C2D14} &\footnotesize C2D\_R50 & \footnotesize 8  & \footnotesize Max &\footnotesize 58.3\% &\footnotesize 58.4\% &\footnotesize 55.8\% &\footnotesize 57.1\% \\ 
 \hline
 
 \footnotesize I3D~\cite{Carreira17} & \footnotesize I3D\_R50\_IN1K & \footnotesize 8  &\footnotesize Max & \footnotesize 55.4\% & \footnotesize 55.5\% & \footnotesize 54.2\% & \footnotesize 54.9\% \\
 \hline

\footnotesize Slow~\cite{Slow21} & \footnotesize Slow4x16 &\footnotesize  4  &\footnotesize Max & \footnotesize  61.1\% & \footnotesize  \textbf{66.7\%} & \footnotesize  57.3\% & \footnotesize  61.6\% \\
 \hline
 
 \footnotesize SlowFast~\cite{SlowFast19} & \footnotesize SlowFast8x8 & \footnotesize 32  &\footnotesize Max & \footnotesize 60.5\% & \footnotesize 60.4\% & \footnotesize 57.7\% & \footnotesize 58.7\% \\
 \hline
 
 \footnotesize NLN~\cite{NonlocalNN17}  &\footnotesize SlowFast8x8 &\footnotesize 8  &\footnotesize Average & \footnotesize  61.6\% & \footnotesize  60.3\% & \footnotesize  58.8\% & \footnotesize  59.5\% \\
 \hline
 
  \footnotesize X3D~\cite{Feichtenhofer20} &\footnotesize X3D\_M &\footnotesize 13 &\footnotesize Max & \footnotesize  60.1\% & \footnotesize  59.1\% & \footnotesize 59.2\% & \footnotesize  59.2\% \\
  \hline
  
  \footnotesize MViTv1~\cite{Fan21} &\footnotesize MViT\_CONV &\footnotesize 32 &\footnotesize Average & \footnotesize  61.8\% & \footnotesize  58.4\% & \footnotesize 58.1\% & \footnotesize  58.2\% \\
  \hline

  \footnotesize MViTv2~\cite{Li21} &\footnotesize MViTv2\_S &\footnotesize 16 &\footnotesize Average & \footnotesize  \textbf{63.9\%} & \footnotesize  64.9\% & \footnotesize \textbf{60.2\%} & \footnotesize  \textbf{62.5\%} \\
 \hline
 \end{tabular}
\caption{\textbf{Performance Comparison of HAR Architectures for Kick Estimation with Two Classes}. This table compares the top backbones used in penalty-kick direction estimation across different architectures, along with the goalkeeper baseline (GK Baseline), corresponding to the actual direction chosen by the goalkeeper during each penalty kick. The table lists the architectures, optimal model variants, number of frames used for HAR embedding extraction, and additional performance metrics.}
  \label{tab:2cls_shoot_zone}
 \end{table*}

Table \ref{tab:3cls_shoot_zone} compares HAR architectures for three-class kick estimation. Each row details an architecture, best model, frame count, pooling technique, and key performance metrics. MViTv1 and MViTv2 achieved the highest accuracy (51.9\% and 51.6\%), excelling in pattern recognition. Architectural choices, such as backbone selection and pooling strategies, significantly impact performance. SlowFast4x16 delivered competitive results, emphasizing tailored architectural design. The choice between max and average pooling influences outcomes, highlighting the need for strategic pooling selection. Table \ref{tab:2cls_shoot_zone} evaluates HAR architectures for binary-class kick estimation (\textit{right}, \textit{left}). Accuracy is generally higher (55.9\% to 63.9\%), with MViTv2 leading at 63.9\%. Slow4x16 achieved 66.7\% precision, while MViTv2 balanced precision (64.9\%), recall (60.2\%), and F1-Score (62.5\%). Pooling strategies significantly impact model performance.

We examined the classification outcomes for both the three-class and two-class tasks through confusion matrix analysis. In the three-class setting, the model correctly classified 60.4\% of left shots and 61.3\% of right shots, while center shots were correctly identified only 12.0\% of the time, with the remaining misclassified as left (46.0\%) or right (42.0\%). This highlights the difficulty of reliably distinguishing the \textit{center} class, likely due to its ambiguous visual definition. From the camera perspective, subtle differences between center and slightly off-center shots are challenging to detect, and the subjective nature of manual annotation may contribute to inconsistencies in labeling. This conceptual vagueness motivated our inclusion of a two-class setup, which removes the ambiguous center category and yields more reliable performance. In this regard, the two-class task showed improved and more stable performance, with left shots classified correctly 64.2\% of the time and right shots 86.5\%, supporting our decision to adopt this simplified setup.

As anticipated, the two-class models consistently outperform the three-class models across both tables, demonstrating higher accuracy. MViTv2 excels in kick direction estimation, achieving 51.6\% accuracy in the three-class task, just behind MViTv1 (51.9\%). However, it leads in binary classification with 63.9\% accuracy, demonstrating its adaptability in simplifying classification while maintaining precision and recall. Its balanced performance—64.9\% precision, 60.2\% recall, and 62.5\% F1-Score—ensures accurate predictions with minimal false positives, making it a reliable choice compared to models like Slow4x16, which, despite higher precision (66.7\%), lacks the same recall and F1-Score.

MViTv2's superior performance stems from its use of average pooling, which provides a holistic view of temporal data rather than emphasizing only salient features like max pooling. This contributes to its ability to capture long-range dependencies and contextual relationships, crucial for analyzing dynamic actions such as penalty kicks. Consequently, transformer-based architectures like MViTv2 excel in interpreting sequential movements, setting them apart from traditional HAR models and solidifying their role in improving kick direction estimation. Metadata incorporation improved accuracy by 2-3\% and F1-Score by 5\%. Using two-stream embedding branches instead of one increased accuracy by 3-5\%, highlighting the importance of integrating running and kicking phases.

\begin{figure}[t]
    \centering
    \begin{minipage}{0.48\textwidth}
        \centering
        \includegraphics[scale=0.45]{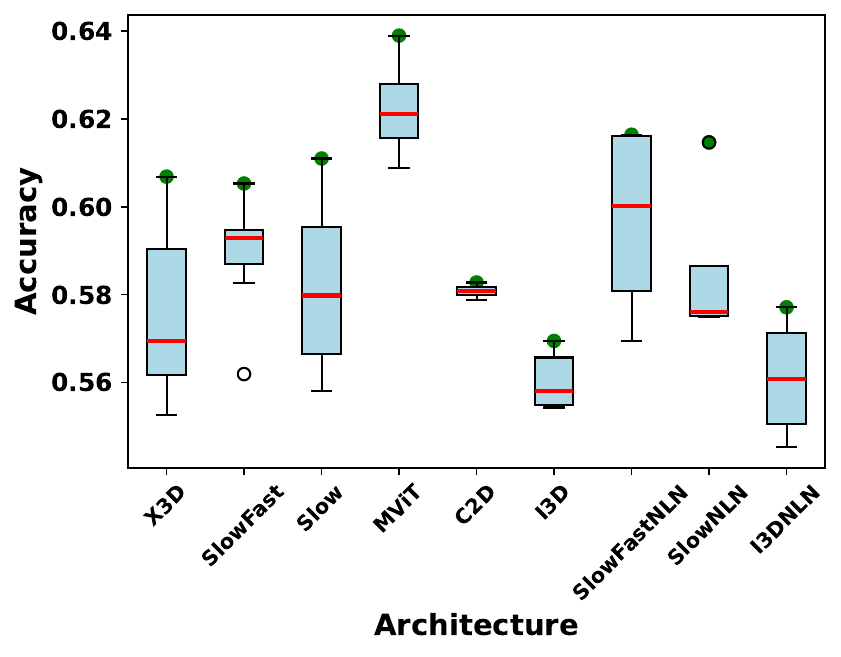}
        \caption{\textbf{Accuracy Analysis.}}
        \label{fig:acc_ana}
    \end{minipage}%
    \hfill
    \begin{minipage}{0.48\textwidth}
        \centering
        \includegraphics[scale=0.45]{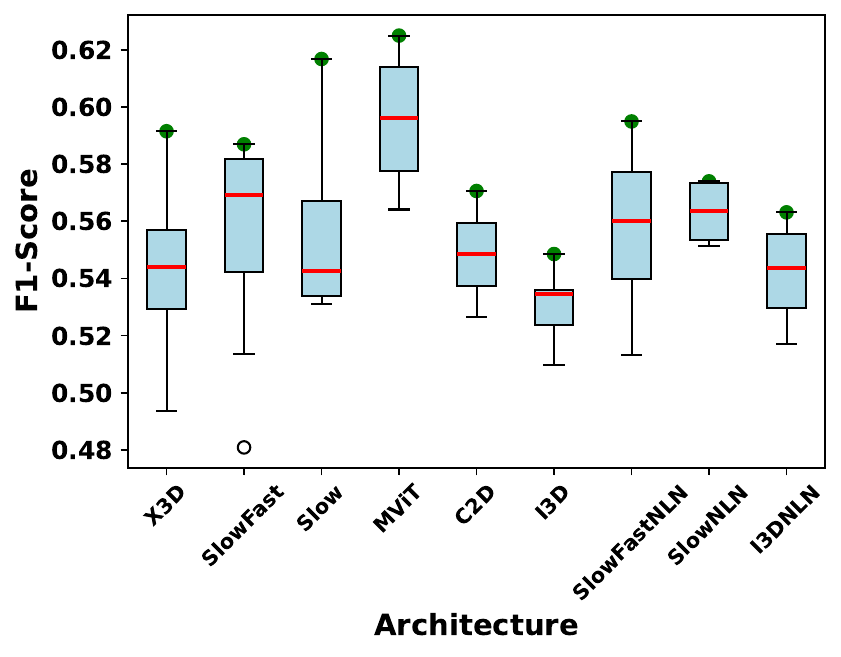}
        \caption{\textbf{F1-Score Analysis.}}
        \label{fig:f1_ana}
    \end{minipage}
\end{figure}

\section{Architecture performance analysis}
\label{sec:ana}

We evaluated 22 HAR models selected to represent a broad range of state-of-the-art architectures. These 22 models correspond to variants across seven core architecture families, offering a representative sample of diverse HAR design strategies. The goal was to cover various HAR design principles and analyze their generalization and robustness in the penalty kick prediction task. While the previous section focused on the best-performing models, this section offers a broader analysis of each architecture's behavior across all its variants. Using boxplots, we highlight performance variability and consistency, helping to identify high-performing models and the most robust and stable architecture families. Figures \ref{fig:acc_ana} and \ref{fig:f1_ana} show the accuracy and F1-Score distributions for all evaluated models under the 2-class experiment configuration. To enhance clarity, NLN variants are grouped according to their source architectures, and green markers indicate the best accuracy achieved by each architecture, aligning with the top-performing instances reported in Table \ref{tab:2cls_shoot_zone}.

Figure \ref{fig:acc_ana} presents a boxplot analysis of accuracy distributions across architectures. MViT models show a broader performance range, with the best-reaching 63.9\% accuracy. In contrast, I3DNLN and Slow models exhibit narrower distributions, indicating more consistent results. SlowFastNLN displays more significant variability, while SlowNLN includes a notable outlier with significantly higher accuracy, suggesting an exceptional configuration. Such outliers may arise from model design or data nuances. Overall, performance trends remain consistent between base models (SlowFast, Slow, I3D) and their NLN-enhanced versions.

Figure \ref{fig:f1_ana} visually captures the variation in F1-Score across the examined HAR architectures. Some architectures exhibit relatively consistent performance, as indicated by narrow interquartile ranges and medians near the center of the boxes. Notably, SlowFast and MViT demonstrate higher median F1-Scores, suggesting more reliable performance across different scenarios. Conversely, architectures like Slow display a wider interquartile range, indicating greater variability in F1-Score performance. This variability may be attributed to the sensitivity of these models to certain aspects of the input data or the complexity of the recognition task.

\section{Conclusions}
\label{sec:con}

This study presents a penalty-kick dataset designed to train deep learning models for short-term human action anticipation. Building on this dataset, we propose a method for predicting penalty-kick direction using HAR embeddings. Our pipeline includes player tracking via ByteTrack, segmentation of running and kicking phases, and feature extraction. We evaluate 22 models across seven architectures—including NLN variants—identifying optimal configurations. Among them, MViTv1 and MViTv2 achieved 51.9\% and 51.6\% accuracy in three-class classification, while MViTv2 led in binary classification with 63.9\%. Precision-recall trade-offs indicate balanced performance.
Significantly, our best-performing models surpass the baseline accuracy of human goalkeepers, highlighting the potential of learned motion cues in decision-making under pressure. These results demonstrate that the dataset contains rich and representative HAR features, enabling effective anticipation of human actions. As such, it offers a valuable benchmark for testing new HAR models and advancing the state of the art. Despite the inherent difficulty of the task, the model's ability to generalize across subtle visual cues supports its practical application in real-time penalty analysis.

\textbf{Acknowledgments}.
This work is partially funded funded by project PID2021-122402OB-C22/MICIU/AEI
/10.13039/501100011033 FEDER, UE and by the ACIISI-Gobierno de Canarias and European FEDER funds under project ULPGC Facilities Net and Grant \mbox{EIS 2021 04}.

%
%
\bibliographystyle{splncs04}

\end{document}